# Semblance: A Rank-Based Kernel on Probability Spaces for Niche Detection

**Authors**

Divyansh Agarwal[1,2], Nancy R. Zhang[1,2]

**Affiliations**

[1]Department of Statistics, The Wharton School, University of Pennsylvania, Philadelphia PA 19104.

[2]Genomics and Computational Biology, Perelman School of Medicine, University of Pennsylvania, Philadelphia PA 19104.

Correspondence:
D.A.: Divyansh.Agarwal@pennmedicine.upenn.edu
N.R.Z.: nzh@wharton.upenn.edu

**Abstract**

In data science, determining proximity between observations is critical to many downstream analyses such as clustering, information retrieval and classification. However, when the underlying structure of the data's probability space is unclear, the function used to compute similarity between data points is often arbitrarily chosen. Here, we present a novel concept of proximity, Semblance, that uses the empirical distribution across all observations to inform the similarity between each pair. The advantage of Semblance lies in its distribution-free formulation and its ability to detect niche features by placing greater emphasis on similarity between observation pairs that fall at the outskirts of the data distribution, as opposed to those that fall towards the center. We prove that Semblance is a valid Mercer kernel, thus allowing it's principled use in kernel-based learning machines. Semblance can be applied to any data modality, and we demonstrate its consistently improved performance against conventional methods through simulations and three real case studies from very different applications: cell-type classification using single-cell RNA sequencing, selecting predictors of positive return on real estate investments, and image compression.

**Introduction**

In modern data analysis, the data are often first reduced to a proximity matrix representing the pair-wise similarity between observations, which becomes the input to downstream analyses such as clustering, classification, and information retrieval. This proximity matrix is an information map as well as an information bottleneck. The former, because all of the information available to the downstream analysis must be extracted from the matrix, and the latter, because the matrix must transmit enough information about the data for any downstream algorithm to be able to do its task. In exploratory data analysis settings, proximity based on Euclidean distance or correlation-based metrics are popular ad hoc choices (*1-3*). More sophisticated, context-specific choices have been designed for particular tasks (*4, 5*).

During the last two decades, efficient kernel-based learning algorithms and their Reproducing Kernel Hilbert Space (RKHS) interpretations have generated intense renewed demand in the development of proximity matrices that satisfy the Mercer condition, which would allow the detection of nonlinear relations using well understood linear algorithms. Such proximity matrices, called Mercer kernels, form the core of several state-of-the-art machine learning systems, and are highly efficacious at solving nonlinear optimization problems (*6, 7*).

Constructing a similarity function amounts to encoding our prior expectation about the possible functions we may be expected to learn in a given feature space, and thus, it is a critical step in real-world data modeling (*8, 9*). Yet, in the initial stages of data analysis, when the underlying structure of the data's probability space is unclear, the choice of the similarity/distance metric is usually arbitrary. Data distributions in the real world are often not elliptical, with continuous and discrete features generally intermixed. At this stage, there is little prior knowledge to guide the selection of distance/similarity measures, much less the design of valid Mercer kernels. Thus, even in this modern age when kernel-based learning algorithms are ripe and sophisticated, we often default to relying on Euclidean distance or Pearson correlation in exploratory data analysis settings.

Here, we develop Semblance, a general, off-the-shelf proximity measure that reveals features from data that are obscured by current commonly-used kernel measures. We prove that Semblance is a valid Mercer kernel, and thus can be used in any kernel-based learning algorithm (*10*). Semblance achieves higher sensitivity for niche features by adapting to the empirical data distribution. Our construction builds on the intuition that in a high dimensional feature space, not all features are equally informative, and furthermore not all values of a given feature should be treated equally. Under independence, it is much more improbable for two objects to agree on a rare feature value than on a common feature value, and thus, two objects agreeing on many rare features suggests that they may belong to a common niche group. Similarly, for a continuous-valued feature, physical proximity in terms of absolute distance is much harder to achieve at the tails of the distribution as compared to at its center. Semblance rewards proximity between objects in the tail of the empirical distribution. By construction, Semblance is nonparametric and invariant to univariate transformations of the features, and therefore is robust.

We first describe the construction of the Semblance function, and then prove that it satisfies Mercer's condition. Then, under simplified but transparent simulation experiments, we systematically explore the types of patterns that we can expect to identify using Semblance. Lastly, through three applications in very distinct fields, one from single-cell biology, one from finance, and another from image analysis, we demonstrate the diverse applicability of the Semblance kernel.

**Constructing the Rank-based Semblance Function**

Suppose we begin with $N_{nxG}$, the data matrix with *n* rows and *G* columns. Let each row correspond to an object, and each column correspond to a feature measured for each object. For ease of notation we let $X = (x_1, \ldots, x_g, \ldots, x_G)$ and $Y = (y_1, \ldots, y_g, \ldots, y_G)$ be two objects, *i.e.*, two rows in the matrix $N_{nxG}$.

Now consider a given feature *g*. Let $\mathbb{P}_g$ be its underlying probability distribution, and, as denoted above, the observed values for this feature in objects *X* and *Y* are $x_g$ and $y_g$, respectively. In practice we don't know $\mathbb{P}_g$, but if we did, we could ask how likely are we, if we were to redraw one of the two values $(x_g, y_g)$, to not land further from the other as the current observed value, while preserving the order between the two. Let $Z$ be the redraw, then this could be expressed as the probability:

$$p_g(x_g, y_g) = p_g(y_g, x_g) := \mathbb{P}_g\{\min(x_g, y_g) \leq Z \leq \max(x_g, y_g)\}. \qquad (1.1)$$

The above probability is a measure of dissimilarity between any two values of feature $g$, see Figure 1. An important but subtle detail is that the probability *includes both endpoints* $x_g$, $y_g$, and therefore $p_g(x_g, x_g) = \mathbb{P}\{x_g\} > 0$. This way of measuring dissimilarity is desirable because it naturally incorporates the information in the underlying probability measure that generated the data. For example, as illustrated in Figure 1, in the binary setting, it is much more rare for two observations to both be equal to 0 if 0 has low probability, and thus the "reward" for $x_g = y_g = 0$ depends on the probability mass at 0. Similarly, in the continuous setting, the reward for proximity between $x_g$ and $y_g$ depends on where the pair falls on the distribution: For the same linear distance between $x_g$ and $y_g$, their dissimilarity is higher when they fall at the center of the distribution than when they fall at the tails.

In practice, $\mathbb{P}_g$ is not known, but with a large enough sample size the empirical distribution $\widehat{\mathbb{P}}_g$ serves as a good approximation, leading to the plug-in empirical estimate $\hat{p}_g(x_g, y_g)$, obtained by substituting $\widehat{\mathbb{P}}_g$ for $\mathbb{P}_g$ in (1.1). This is reminiscent of empirical Bayes methods, where information is borrowed across all observed values to inform our dissimilarity evaluation between any given pair. We define:

$$k_g(X, Y) = 1 - \hat{p}_g(x_g, y_g)$$
$$= \frac{1}{n} \sum_{i=1}^{n} [1 - I(\min(x_g, y_g) \leq N_{ig} \leq \max(x_g, y_g))],$$

the empirical probability of falling *strictly* outside the interval $[x_g, y_g]$. Suppose feature $g$ is continuous, and hence each observed value is unique, and let $r_X$, $r_Y$ be the ranks of $x_g$, $y_g$ among all observed values of this feature across the $n$ objects. Then it follows that:

$$k_g(X, Y) = \frac{1}{n}(|r_X - r_Y| + 1).$$

For discrete features, the computation of $k_g(X, Y)$ is more complicated due to ties. Nevertheless, computation of $k_g(X, Y)$ in general is easy and fast. An example algorithm is provided in Methods.

We now define the Semblance function $K(X, Y)$ as simply the mean of $k_g(X, Y)$ across $g$:

$$K(X, Y) = \frac{1}{G} \sum_{g=1}^{G} k_g(X, Y), \qquad (1.2)$$

Since $\hat{p}_g(x_g, x_g) \leq \hat{p}_g(x_g, y_g)$ if $x_g \neq y_g$, it follows that $K(X, X) \geq K(X, Y) \ \forall \ X \neq Y$. Thus, when applied to any data matrix $N$, this function outputs a symmetric $n \times n$ matrix whose rows and columns are maximized at the diagonal.

## Results

**Semblance is a valid Mercer kernel**

Since $K(X, Y)$ is just the mean of $k_g(X, Y)$ across $g$, we start by considering
$$K_g = \{K_g(i,j) = k_g(N_{ig}, N_{jg}) : 1 \leq i, j \leq n\},$$
the matrix derived only from observations of feature $g$. First assume that the objects have been permuted such that $\{N_{ig} : i = 1,\ldots,n\}$ are monotone nondecreasing. Define:
$$a_i = \widehat{\mathbb{P}}(Z < N_{ig}) \text{ and } b_i = \widehat{\mathbb{P}}(Z > N_{jg}), \tag{1.3}$$
suppressing the notational dependence of $a_i$ and $b_i$ on $g$, for simplicity. Based on (1.1), for $i \leq j$,
$$K_g(i,j) = a_i + b_j \tag{1.4}$$
By our monotone nondecreasing assumption, $a_i \leq a_{i+1}$ and $b_i \geq b_{i+1}$. Thus, $K_g$ has the decomposition:

$$K_g = \begin{bmatrix} a_1 & a_1 & \cdots & a_1 \\ a_1 & a_2 & \cdots & a_2 \\ \vdots & \ddots & & \vdots \\ \vdots & & \ddots & \vdots \\ a_1 & a_2 & \cdots & a_n \end{bmatrix} + \begin{bmatrix} b_1 & b_2 & \cdots & b_n \\ b_2 & b_2 & \cdots & b_n \\ b_3 & b_3 & \cdots & b_n \\ \vdots & & \ddots & \vdots \\ b_n & b_n & \cdots & b_n \end{bmatrix} = M + N.$$

**Remark:** The matrices $M$ and $N$ have a symmetric and analogous structure. The left-upper hook comprising the first row and column of $M$ has all entries $a_1$, the second hook has all entries $a_2$ and so on, until the $n^{th}$ hook which is simply the entry $a_n$. Similarly, the right-lower hook of $N$ comprising the last row and column has all entries $b_n$, all the way up to the solo entry $b_1$ in the first row and column.

*Proposition 1: $M$ is a nonnegative-definite (NND) matrix*
*The Proof is by induction.* For the base case, consider the $2 \times 2$ matrix
$$\dot{M} = \begin{bmatrix} a_1 & a_1 \\ a_1 & a_2 \end{bmatrix}$$
By construction $a_{i+1} \geq a_i$, therefore $det(\dot{M}) \geq 0$, and hence $\dot{M}$ is NND. The induction hypothesis is that all $m \times m$ matrices, $\bar{M}$, with the structure

$$\begin{bmatrix} a_2 & a_2 & \cdots & a_2 \\ a_2 & a_3 & \cdots & a_3 \\ a_2 & a_3 & \cdots & a_4 \\ \vdots & & \ddots & \vdots \\ a_2 & a_3 & \cdots & a_m \end{bmatrix} \text{ where } a_2 \leq a_3 \leq \ldots \leq a_m, \tag{1.5}$$

are NND. Now, to prove that the same is true for $m \times m$ matrices, we can write such matrices in the form

$$\begin{bmatrix} a_1 & U \\ U^T & \bar{M} \end{bmatrix}, \tag{1.6}$$

where $U$ represents the vector $(a_1 a_1 a_1 \cdots)$ and $\bar{M}$ is a matrix that satisfies the induction hypothesis. Using the Schur complement condition for the nonnegative definiteness of a symmetric matrix (*11*), we can show that $\bar{M} - U^T a_1^{-1} U$ is NND:

$$\bar{M} - U^T a_1^{-1} U = \bar{M} - \begin{pmatrix} a_1 \\ a_1 \\ \vdots \\ a_1 \end{pmatrix} \frac{1}{a_1} (a_1 a_1 a_1 \cdots)$$

$$= \bar{M} - \begin{bmatrix} a_1 & a_1 & \cdots & a_1 \\ a_1 & a_1 & \cdots & a_1 \\ \vdots & & \ddots & \vdots \\ a_1 & a_1 & \cdots & a_1 \end{bmatrix}$$

$$= \begin{bmatrix} a_2 - a_1 & a_2 - a_1 & \cdots & a_2 - a_1 \\ a_2 - a_1 & a_3 - a_1 & \cdots & a_3 - a_1 \\ \vdots & & \ddots & \vdots \\ \vdots & & & \vdots \\ a_2 - a_1 & a_3 - a_1 & \cdots & a_n - a_1 \end{bmatrix}$$

This resultant matrix is of a form that satisfies (1.5) and thus, by the induction hypothesis, is NND. Therefore the matrix (1.6) is also NND.

Since, by construction, $N$ mirrors the properties of $M$, we have by *Proposition 1* that $N$ is also an NND matrix. To facilitate our discussion of technical details, we review two facts about NND matrices: (i) the sum of NND matrices is NND, and (ii) permuting the observations of an NND matrix preserves the NND structure. Corresponding proofs for these are provided in Supplementary Methods.

Based on these facts, together with *Proposition 1,* the kernel matrix $K$ (sum of all $K_g$'s) is NND. The matrix $K$ computed on any data matrix by the Semblance function defined in (1.2) is

NND, and thus Semblance is a valid Mercer kernel. As a result, the Representer theorem allows effective implementation of nonlinear mappings through inner products represented by our kernel function (*6, 10*).

**Semblance is conceptually different from rank-based similarity measures**

Since, in the case where all features are continuous, Semblance can be simplified to a function on ranks, we first clarify how it differs from existing rank-based similarity measures: Spearman's Rho ($\rho$) and Kendall's Tau ($\tau$). By construction, Semblance is fundamentally different from these existing measures in two ways. First, while $\rho$ and $\tau$ are based on ranks computed by ordering the values *within each object* (the rows of matrix *N*), Semblance is computed using ranks determined by ordering the values *within each feature* (the columns of matrix *N*). Thus, the Semblance kernel can be expected to produce values that differ substantially from these two measures. Second, Semblance treats ties differently from simple rank-based methods, such that ties shared by many objects diminish the proximity between those objects. This treatment of ties, for discrete data, makes Semblance more sensitive for niche subgroups in the data. Therefore, Semblance is better understood through the lens of empirical Bayes, where, for each feature, the empirical distribution across all objects informs our evaluation of the similarity between each pair of objects.

Simulations allow us to compare the effectiveness of similarity/distance measures under simplified but interpretable settings. We used simulations to compare Semblance against Euclidean distance, Pearson correlation, and Spearman correlation in their ability to separate two groups in an unsupervised setting. We simulated from a two group model, where multivariate objects either came from group 1, with probability $q < 0.5$, or from group 2, with probability $1-q$. Let each object contain *m* features, drawn independently, with a proportion $p \in (0, 1)$ of the features being informative. The informative features have distribution $P_{I,1}$ in group 1 and $P_{I,2}$ in group 2. The rest of the features are non-informative, and have the same distribution $P_{NI}$ across both groups. We consider both continuous and discrete distributions for the features. In the continuous case, the features are generated from:

$$P_{NI} = N(0,1), \quad P_{I,1} = N(\mu\sigma_2,\sigma_1), \quad P_{I,2} = N(0,\sigma_2). \tag{1.7}$$

In the discrete case, the features are generated from:

$$P_{NI} = P_{I,2} = Bernoulli(r_0), \quad P_{I,1} = Bernoulli(r_1). \tag{1.8}$$

Of course, whether a feature is informative or not, and whether an object is from group 1 or group 2, is not used when computing the similarity/distance matrix.

As shown in Figure 2A, in each simulation run, we generated *n* objects with the first $n_1 = qn$ coming from group 1 and the next $n_2 = (1 - q)n$ coming from group 2. Our goal is to detect the existence of the minority group 1 and assign objects to the appropriate group. Similarities (Semblance, Pearson, Spearman) and distances (Euclidean) are computed on this data, each producing an $n \times n$ matrix, which we will call *S*. Let:

$$\bar{S}_{11} = \frac{1}{n_1} \sum_{1 \le i < j \le n_1} S_{ij}, \quad \bar{S}_{22} = \frac{1}{n_2} \sum_{n_1 < i < j \le n} S_{ij}, \quad \bar{S}_{12} = \frac{1}{n_1 n_2} \sum_{1 \le i \le n_1 < j \le n_2} S_{ij}.$$

Then $\bar{S}_{11}$ is the mean similarity/distance between objects in group 1, $\bar{S}_{22}$ is the mean similarity/distance between objects in group 2, and $\bar{S}_{12}$ is the mean similarity/distance across groups. To quantify the signal in $S$, we let $T_1 = (\bar{S}_{11} - \bar{S}_{12})/\text{se}_1, T_2 = (\bar{S}_{22} - \bar{S}_{12})/\text{se}_2$, where se$_1$, se$_2$ are standard errors of the differences in the numerators. Hence, large positive values of $T_1$, $T_2$ imply that downstream algorithms based on $S$ will be able to separate the two groups well.

Figure 2B shows the $T_1$ and $T_2$ values for an example set of simulations where $n=m=100$, the proportion of informative features is 10%, the rare subpopulation proportion is 10%, and every feature is normal following (1.7) with μ = 2 and $\sigma_1$, $\sigma_2$ varying from 0.1 to 1. Heatmaps in the top row show the values of $T_1$ and those in the bottom row show the values of $T_2$ for each of the four similarity/distance measures. We see that Semblance improves upon Euclidean distance, Pearson, and Spearman, attaining large values for $T_1$ and $T_2$ across a broad range of parameters, especially when $\sigma_2$ is small. Figure 2C shows another set of simulations, with the same $n$, $m$ and $p$ values as Figure 2B, but under the model (1.8) with $r_2 = 0.5$, $r_1$ varying from 0.01 to 0.2, and $q$ varying from 0.05 to 0.5. We see that in this case, there is no signal in $T_2$ for all of the measures except Semblance, and in fact, both Pearson and Spearman correlation fail to separate the two groups for much of the parameter range. In contrast, Semblance gives large values for both $T_1$ and $T_2$ for a large portion of the explored parameter region.

We explored varying combinations of $p$, $q$, $\sigma_1$ and $\sigma_2$ in the normal setting, and $p$, $q$, $r_0$ and $r_1$ in the Bernoulli setting. Summarizing these systematic experiments in representative heatmaps (Fig. 3), we found that Semblance has robust performance across different distributions and distribution parameters ($\sigma_1$, $\sigma_2$, $r_1$, $r_2$) as long as the proportion of informative features is not too small. Semblance is better than the other metrics especially in differentiating small tight subpopulations, *i.e.*, niche groups. Unweighted Semblance ($w_g =1$ in (1.2)) retains less information and should not be used when informative features are extremely rare (p→0) but the separation between clusters is extremely large (p→0, μ→∞).

**Semblance kernel-tSNE identifies a niche retinal horizontal cell population**

In the setting of single-cell RNA sequencing (scRNA-seq), the data is in the form of a matrix with each row representing a cell, and each column representing a gene. For cell $c$ and gene $g$, N$_{cg}$ is a count matrix measuring a gene's RNA expression level in the given cell. A first step in the analysis of such data is often visualization via a t-distributed Stochastic Neighbor Embedding (tSNE)-type dimension reduction. Most studies arbitrarily use the Euclidean distance or RBF in this step, although methods based on more sophisticated kernel choices that rely on strong prior assumptions have been proposed (*12*). Starting from the low-dimension embedding, a primary goal in many single-cell studies is to classify cells into distinct cell-types and identify previously unknown cell subpopulations. This is a challenging analysis due to many factors: (1) Expression levels are not comparable across genes, lowly expressed cell type markers may be swamped by highly expressed housekeeping genes; (2) gene expression at the single cell level is often bursty and thus cannot be approximated by the normal distribution; (3) one is often interested in detecting rare niche subpopulations for which current methods have low power. These considerations motivate the use of the Semblance kernel to compute a cell-to-cell similarity metric, which can be used as input to tSNE, kPCA, and other kernel-based algorithms. Most methods used for cell-type identification based on scRNA-seq limit their consideration to highly variable genes, thereby using only a subset of the features. Instead, Semblance can be computed over all features, ensuring that information from all informative genes is retained.

Consider Retinal Horizontal Cell (RHC), a unique cell-type that recently came to limelight due to its notable morphological plasticity, and its role as the possible precursor for retinoblastoma (*13*). RHCs have a special level of complexity wherein they can undergo migration, mitosis and differentiation at late developmental stages. They are traditionally divided into H1 axon-bearing and H2-H4 axon-less subtypes, although the latter are largely absent in the rod-dominated retina of most mammals (*14*). The axon-bearing and axon-less RHC subtypes are generated during retinal development from progenitors that are susceptible to a transition in metabolic activity. For example, Follistatin, an anabolic agent that alters protein synthesis and the inherent metabolic architecture in tissues, increases RHC proliferation (*15*). RHC subtypes also exhibit temporally distinguishable periods of migration, likely affected by their cellular metabolic state. These distinctive features are controlled by a niche set of genes, and thus RHCs provide a nonpareil setting to test Semblance. We employed our kernel on an scRNA-seq dataset of 710 Lhx1$^+$ RHCs from healthy P14 mouse (*16*), and sought to answer the question: how similar are RHCs to each other? When we use Euclidean distance for tSNE analysis, only one RHC cluster could be identified (Fig. 4A), as opposed to two subsets of RHCs identified using kernel-tSNE (Fig. 4B). Furthermore, when we mapped the two clusters obtained using Semblance back to the Euclidean distance-based tSNE projection, we found that Semblance led to a better visual separation between the two clusters, which were otherwise harder to distinguish from each other (Fig. S1). We then sought further biological interpretation of these results and discovered that the cells in the second, smaller cluster − comprising 12% of the total RHC population − identified by Semblance exhibit differential expression of genes and pathways that affect metabolism (Fig. 4C). We explicated our results by testing for enriched Gene Ontology (GO) functional categories using REVIGO (*17*), and uncovered a niche RHC population that has unique metabolic response properties (Fig. 4D).

**An kernel SVM approach using Semblance is comparable to other kernels at predicting stock market returns for real estate investment companies**

In finance and business analytics, stock market forecasting remains an active area of research. Although predictions of market volatility are inherently challenging, support vectors machines (SVMs) using a Gaussian or Laplacian kernel have been found to be efficient and accurate in modeling stock market prices, partly because training an SVM requires convex optimization with a linear constraint, which often has a stable, unique global minimum (*18, 19*). Semblance is well poised to detect niche features, and thus we hypothesized that this attribute would make Semblance useful in kernel-based classification problems. To examine this proposition, we used the Center for Research in Security Prices (CRSP) Database (http://www.crsp.com/products/research-products/crspziman-real-estate-database), which combines stock price and returns data with financial indices and company-specific information on all real estate investment trusts (REITs) that have traded on the three primary exchanges: NASDAQ, NYSE and NYSE MKT. We focused our analysis on the 5,419 REITs that were actively trading between January 1, 2016 and December 31, 2017, and obtained a list of financial indices and company-specific indicators for each REIT (Table S1).

We determined which REITs had a net positive rate of return on their stock, and then classified the companies based on whether they had a positive or negative rate of return. In accordance with previous methods (*20, 21*), we used kernel SVM (kSVM) to determine how accurately the model was able to predict the REIT category. We randomly split the REIT data into training and test data in a 3:1 ratio, and compared the performance of Semblance against eight other kernel functions. To compare the generalization ability of each kSVM classifier, we used a 10-fold cross validation to estimate the true test accuracy as it averaged over ten runs. Although further improvement in SVM performance results can be obtained possibly by incorporating more financial indices, and REIT characteristics, our results are empirically consistent with previous research in this area. We found that Semblance was more accurate at REIT classification than majority of the other kernel choices (Table S2), and also observe that the Gaussian, laplacian and polynomial kernels perform better than linear, spline and hyperbolic tangent kernels in this scenario, likely because the former set of kernels are homogenous and have good approximation capabilities. Notably, this example demonstrates that the performance of Semblance kSVM is comparable to other popularly used kernels, such as the radial basis kernel.

**Semblance kernel PCA is efficient at image reconstruction and compression**

Kernel Principal Component Analysis (kPCA), the nonlinear version of PCA, exploits the structure of high-dimensional features, and can be used for data denoising, compression and reconstruction (*22, 23*). This task, however, is nontrivial because the kPCA output resides in some high-dimensional feature space, and does not necessarily have pre-images in the input space (*24*). kPCA, particularly using the Gaussian kernel defined by $k(x,y) = \exp(-||x-y||^2/2\sigma^2)$, has been used extensively to improve active shape models (ASMs), reconstruct pre-images and recreate compressed shapes due to its ability to recognize more nuanced features in real-world pictures (*25*). Nonlinear demunging and data recreation based on kPCA rests on the principle that using a small set of some *f* kPCA features provides an *f*−dimensional re-parametrization of the data that better captures its inherent complexity (*26*). Since Semblance is rank-based and emphasizes rare or niche subgroups in data, that led us to surmise that it would be effective as a nonlinear image denoising and reconstruction method. We discovered that Semblance kPCA can indeed be used to reconstruct real-world images with remarkably good performance (Fig. 5A and B). Upon adding uniform noise to an image, we found that Semblance kPCA is particularly useful for image de-noising and compares favorably against linear PCA and Gaussian kPCA (Fig. 5C and D).

We further evaluated the performance of kPCA on pictures obtained from The Yale Face Database (http://cvc.cs.yale.edu/cvc/projects/yalefaces/yalefaces.html) and the Bioconductor package EBImage (*27*), and found that Semblance can give a good re-encoding of the data when it lies along a nonlinear manifold, as is often the case with images. In each experiment, we computed the projections of the given image data onto the first *f* components and then sought to reconstruct the image as precisely as possible. We found that Semblance kPCA performed better than linear PCA and Gaussian kPCA when using a comparable number of components (Fig. S2). This encouraging observation is supported by the intuitions underlying the construction of Semblance. Linear PCA encapsulates the coarse data structure as well as the noise. In contrast, Gaussian kPCA, similar to a k-nearest neighbor method, apprehends the connection between data points that are close to each other in the original feature space (*28*). On the other hand, Semblance is based on ranks and not absolute values, and therefore performs better when informative and non-informative features in an image are not on the same scale.

**Discussion**

Here, we present a novel rank-based semblance kernel on probability spaces that is useful for constructing a similarity matrix, and is particularly powerful at detecting niche features. Semblance operates in a high-dimensional, implicit feature space and can be applied to any data domain. We have shown that Semblance is a valid Mercer kernel and thus it provides a principled approach for detecting nonlinear relations using well-understood linear learning algorithms. This is further substantiated by its performance is two very diverse applications in completely independent fields of study. From a computational point-of-view, Semblance enables the extraction of features of the data's empirical distribution at low computational cost. The use of feature ranks on a probability space ensures that Semblance is robust to outliers and statistically stable, therefore making it a widely applicable algorithm for pattern analysis. Our kernel method has attractive unbiasedness and power compared to existing, commonly-used similarity measures, as shown through simulations and real-data examples. Moreover, Semblance will also find tremendous utility in "multiple kernel learning" approaches wherein multiple kernels are often combined to learn from a heterogeneous data source.

**Materials and Methods**

**Algorithm to implement the Semblance kernel**

---

**procedure** STEP 1
    For a given feature, $g$, create a descending ranked list
    such that the object with the highest value of $g$
    is ranked 1, and the object with lowest value is
    ranked last.
**procedure** STEP 2
    Compute the empirical CDF for the feature $g$.
    *loop*:
    **for** features $g : 1 \to G$ **do**
    Store lists as look-up tables
    For a given feature $g$, determine where two
    observations, $x$ and $y$, fall on the CDF for $g$.
**procedure** STEP 3
    Calculate the difference between the ranks of $x$ and $y$,
    Add 1 to the difference between ranks.
**procedure** STEP 4
    *loop*:
    **for** features $g : 1 \to G$ **do**
    *Step 3*, and store the cumulative sum in a matrix as
    entry $(x, y)$, corresponding to the $x^{th}$ row and $y^{th}$ column

---

***R* package implementation**

Semblance is an open-source R package available on CRAN (https://cran.r-project.org/web/packages/Semblance/), and is compatible with existing kernel method libraries such as kernlab (*29*). In our R package, we implemented the kernel method in the *ranksem* function, which takes an input $N_{ng}$ matrix (of $g$ feature measurements for $n$ objects), and returns an $n \times n$ similarity matrix.

**Proofs concerning nonnegative definite matrices**

*Lemma 1:* The sum of NND matrices is NND.
*Proof:* Let $K_{g(1)}$ and $K_{g(2)}$ be two NND matrices, such that $\forall z \in \mathbb{R}^n$ :

$$z^T K_{g(1)} z \text{ and } z^T K_{g(2)} z > 0 \Rightarrow z^T K_{g(1)} z + z^T K_{g(2)} z > 0$$

Using the distributive law of matrix multiplication:

$$0 < z^T K_{g(1)} z + z^T K_{g(2)} z = z^T \left( K_{g(1)} + K_{g(2)} \right) z$$
$$\Rightarrow z^T \left( K_{g(1)} + K_{g(2)} \right) z > 0 \therefore \left( K_{g(1)} + K_{g(2)} \right) \succ 0.$$

*Lemma 2:* Permuting the observations of an NND matrix preserves the NND structure.
*Proof:* Let π be the permutation matrix such that it has exactly one entry in each row and in each column equal to 1, and all other entries are 0. For any permutation matrix, $\pi^{-1} = \pi^T$ and thus:
$$\pi \pi^T = \pi^T \pi = I$$
For any given NND matrix, $K$, $\pi K \pi^T$ is also NND. Clearly, $\pi K \pi^T$ is also symmetric as:
$$w^T (\pi K \pi^T) w = (\pi^T w)^T K (\pi^T w) \quad \forall w \neq 0 \quad \text{since } K \text{ is NND}$$

Furthermore, every NND matrix can be factored as $K = A^T A$, where $A$ is the Cholesky factor of $K$. The Cholesky factorization of NND matrices is numerically stable – a principal permutation of the rows and columns does not numerically destabilize the factorization (*30*). This leads to the result that symmetrically permuting the rows and columns of an NND matrix yields another NND matrix.

## Acknowledgments

N.R.Z would like to thank Dr. Zhijin Wu (Brown University, Rhode Island) for enlightening discussions. We are also grateful to Dr. Hua Tang (Stanford University, California) for providing feedback on an earlier version of this research report.

**Funding:** This work was supported by the National Institutes of Health (NIH grant R01 HG006137 to D.A. and N.R.Z.).

**Author contributions:** D.A. and N.R.Z. devised the idea, conducted the supporting experiments and wrote the manuscript.

**Competing interests:** The authors have no conflicts of interest.

**Data and materials availability:**
All data needed to evaluate the conclusions in the paper are described in the paper and/or the Supplementary Materials, and are publicly available on open-access repositories. Any additional computer codes related to this paper may be requested from the authors.
The scRNA-seq data were generated using the Drop-seq platform, and are publicly accessible via the NCBI Gene Expression Omnibus (Accession: GSE63473). The data on Real Estate Investment Trusts (REITs) is curated by the Center for Research in Security Prices (CRSP), and is accessible at: http://www.crsp.com/products/research-products/crspziman-real-estate-database

**Figures and Tables**

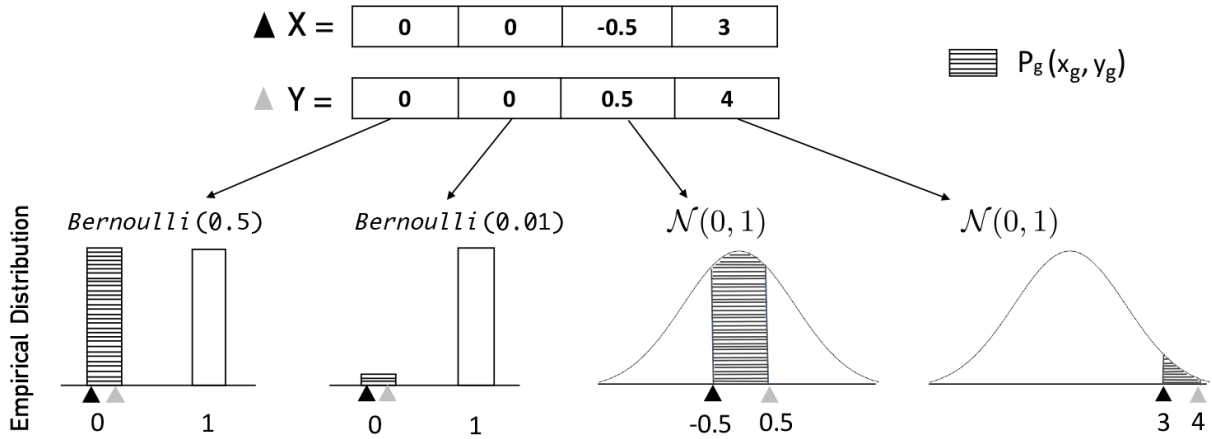

**Fig. 1. Illustration of what $p_g(x_g, y_g)$ corresponds to in the case of a discrete distribution or a continuous distribution.** In this toy example, $X$ and $Y$ are two objects with four features measured. Semblance computes an empirical distribution from the data for each feature, and uses the information of where the observations fall on that distribution to determine how similar they are to each other. Specifically, it emphasizes relationships that are less likely to occur by chance and that lie at the tail ends of a probability distribution. For example, $X$ and $Y$ are equal to 0 for both the first and second feature, but these two features contribute different values to the kernel: "0" is more rare for the second feature, and thus $p_2(0, 0)$ is smaller than $p_1(0, 0)$ and the second feature contributes a higher value in the Semblance kernel. Similarly, even though the difference between $X$ and $Y$ is 1 for both features 3 and 4, feature 4, where the values fall in the tail, has lower $p_g(x_g, y_g)$ and thus contributes a higher value in the Semblance kernel than feature 3.

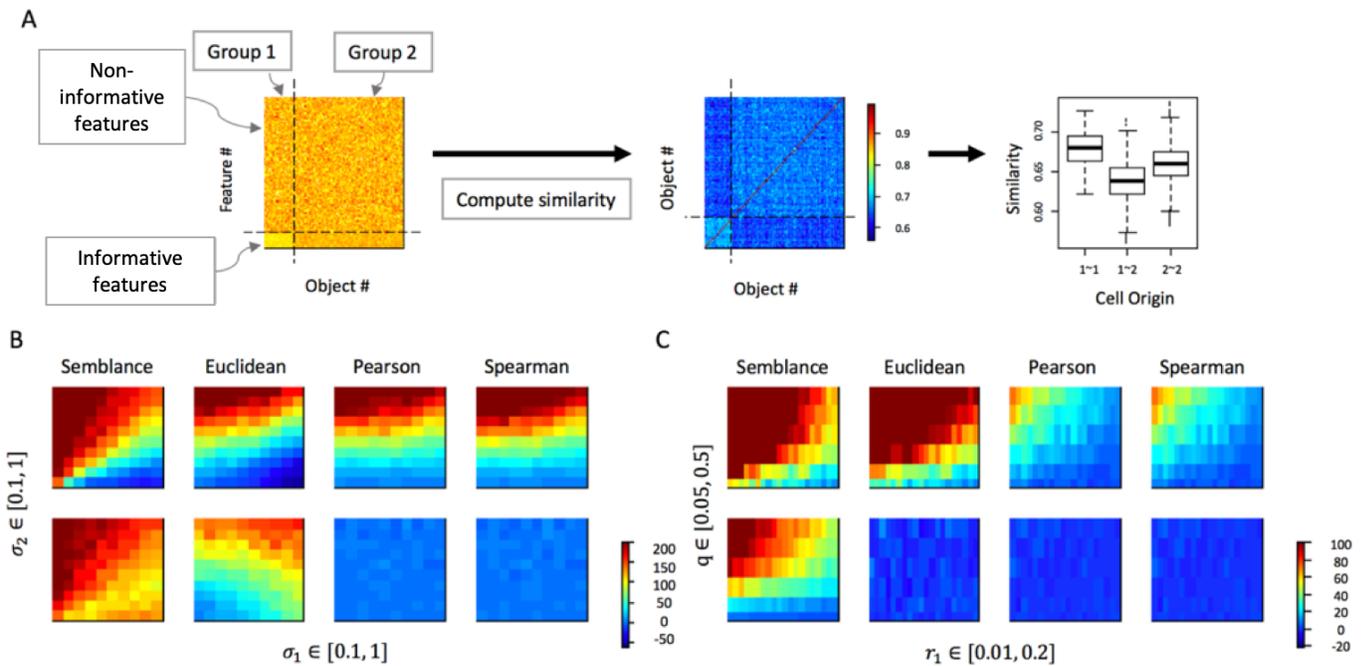

**Fig. 2. Simulations exploring the effectiveness of similarity/distance measures. (A)** Set-up for one simulation run. **(B)** $T_1$ (top) and $T_2$ (bottom) values for each similarity/distance metric,

for varying values of $\sigma_1 \in [0.1, 1]$ (horizontal axis) and $\sigma_2 \in [0.1, 1]$ (vertical axis). **(C)** $T_1$ (top) and $T_2$ (bottom) values for each similarity/distance metric, for varying values of $r_1 \in [0.1, 1]$ (horizontal axis) and $q \in [0.1, 1]$ (vertical axis).

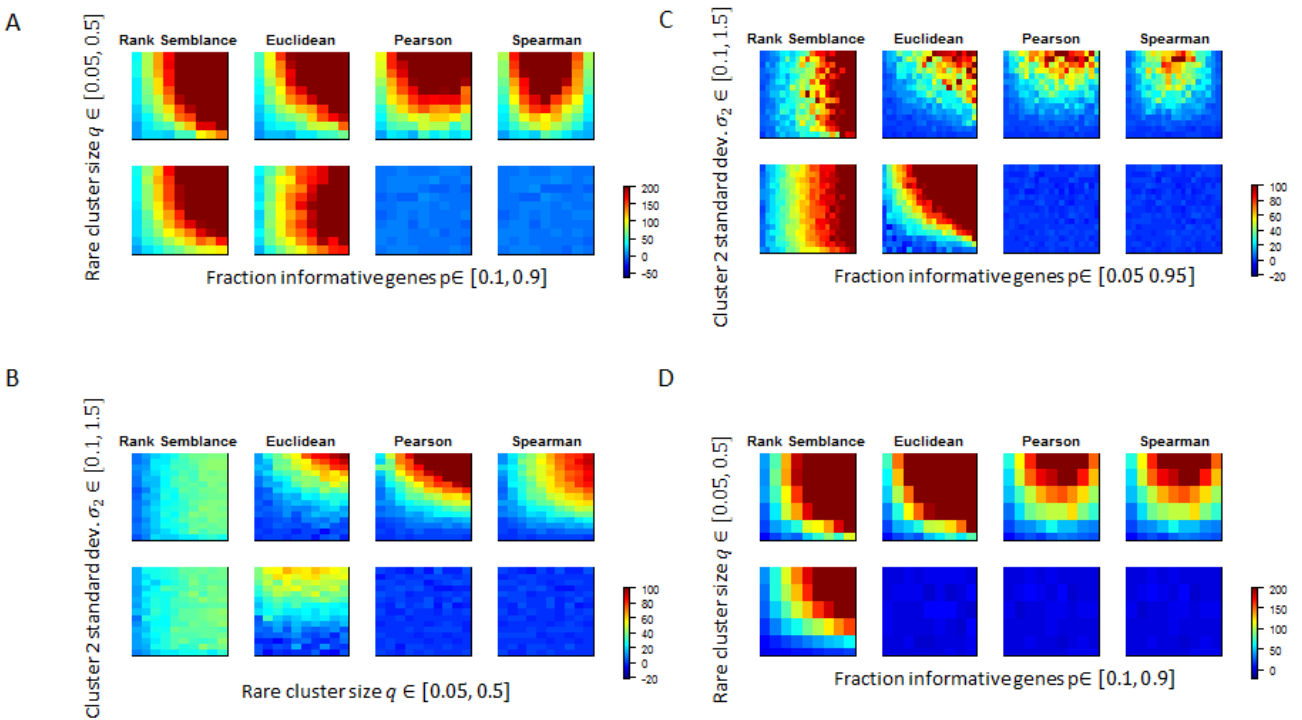

**Fig. 3. Simulation results over parameter sweeps.** For each 2 by 4 group of heatmaps, the top row shows $T_1$ and the bottom row shows $T_2$ for each similarity/distance metric, computed as described in the text. Simulation parameters are varied along the rows and columns of the heatmaps. **(A)** Normal model, $p = \{0.1, 0.2, \ldots, 0.9\}$ for horizontal axis, and $q \in \{0.05, 0.1, \ldots, 0.5\}$ for vertical axis. **(B)** Normal model, $q \in \{0.05, 0.1, \ldots, 0.5\}$ for horizontal axis and $\sigma_2 \in \{0.1, 0.2, \ldots, 1.5\}$ for vertical axis. **(C)** Normal model, $p = \{0.1, 0.2, \ldots, 0.9\}$ for horizontal axis and $\sigma_2 \in \{0.1, 0.2, \ldots, 1.5\}$ for vertical axis. **(D)** Binomial model, $p = \{0.1, 0.2, \ldots, 0.9\}$ for horizontal axis, and $q \in \{0.05, 0.1, \ldots, 0.5\}$ for vertical axis.

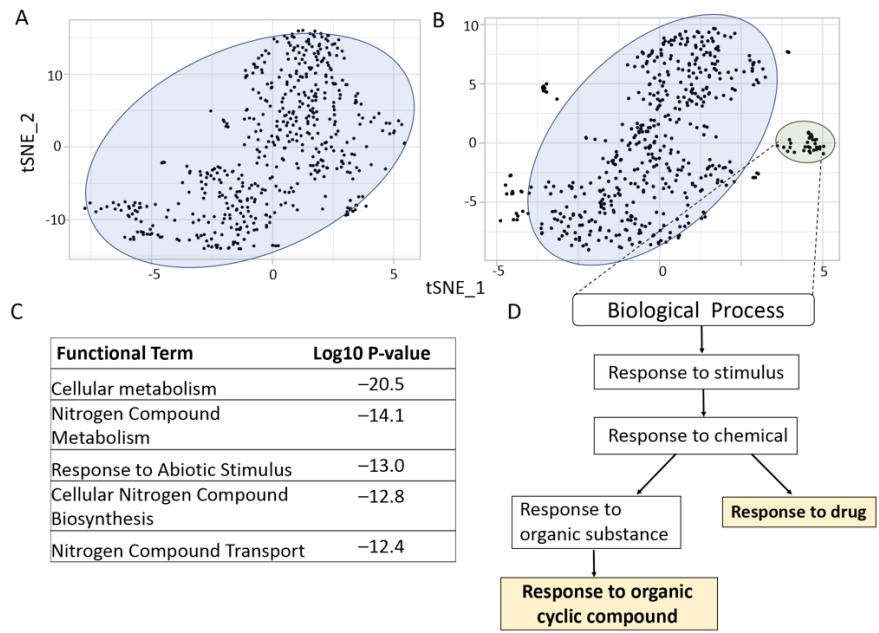

**Fig. 4. A unique RHC cluster is identified by Semblance k-tNSE.** Each black dot in **(A-B)** represents a single cell. Euclidean distance tSNE identifies a single RHC cluster **(A)** as opposed to two sub-populations identified by Semblance **(B)**. The top 5 pathways found to be enriched in this cellular subtype are shown **(C)**, and GO Analysis suggested that the smaller cluster has unique metabolic response properties **(D)**.

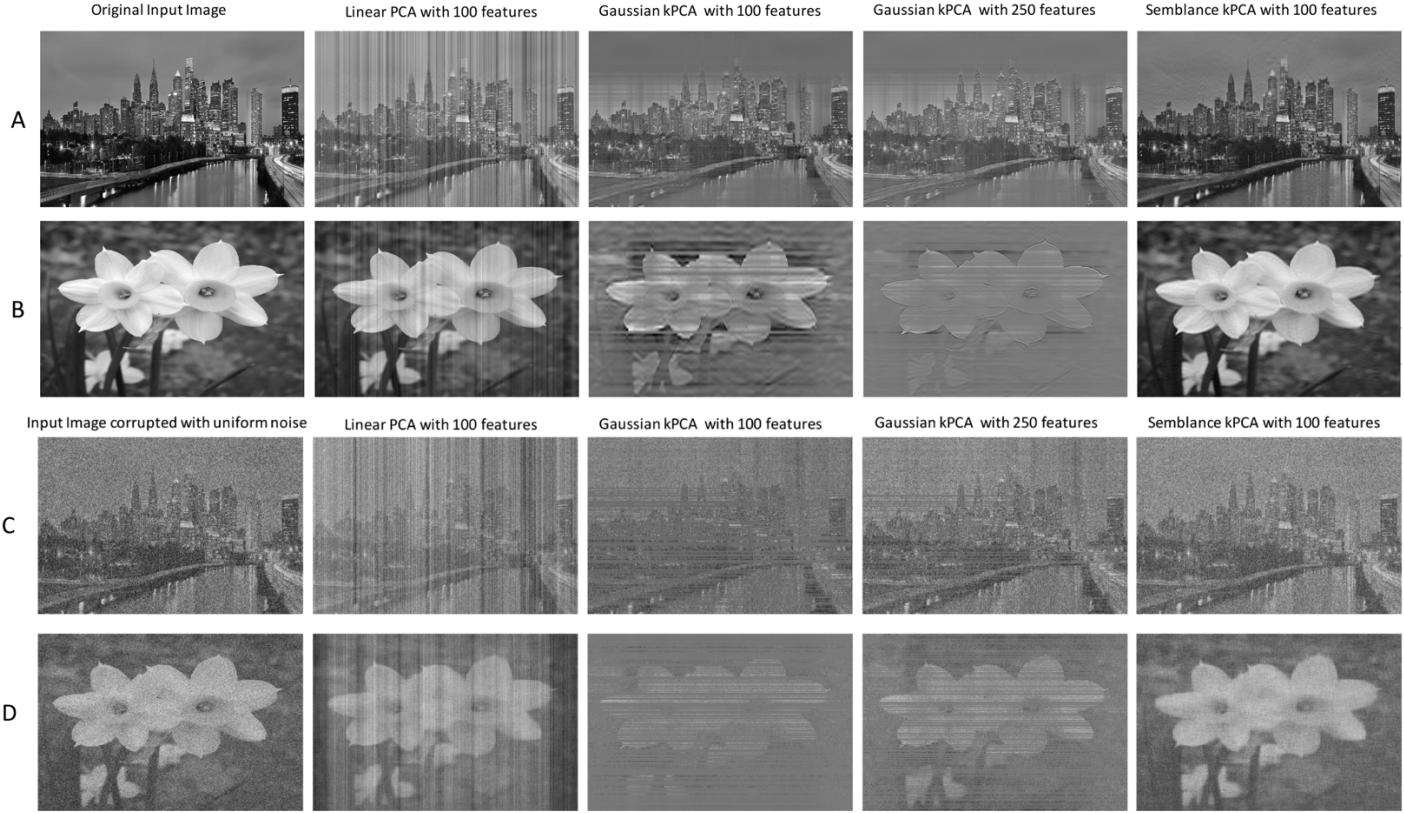

**Fig. 5. kPCA using the Semblance kernel provides a useful method for image reconstruction and denoising.** Two example open-source images: Philadelphia skyline **(A)**, and Daffodil flowers **(B)** are shown here. Semblance kPCA was able to effectively recover and compress when compared with linear PCA or Gaussian kPCA. These images were corrupted with added uniform noise: **(C)** and **(D)**, respectively. The recovered image output using linear PCA, Gaussian kPCA and Semblance kPCA is displayed. Comparing the same number of features (and even 2.5x as many features for Gaussian kPCA), Semblance performs favorably. More examples are given in the supplement. *Photo Credits*: Mo Huang (The Wharton School) and the EB Image Package.

**Supplementary Material**

**Proofs concerning nonnegative definite matrices**

*Lemma 1:* The sum of NND matrices is NND.

*Proof:* Let $K_{g(1)}$ and $K_{g(2)}$ be two NND matrices, such that $\forall z \in \mathbb{R}^n$ :

$$z^T K_{g(1)} z \text{ and } z^T K_{g(2)} z > 0 \Rightarrow z^T K_{g(1)} z + z^T K_{g(2)} z > 0$$

Using the distributive law of matrix multiplication:

$$0 < z^T K_{g(1)} z + z^T K_{g(2)} z = z^T \left( K_{g(1)} + K_{g(2)} \right) z$$

$$\Rightarrow z^T \left( K_{g(1)} + K_{g(2)} \right) z > 0 \therefore \left( K_{g(1)} + K_{g(2)} \right) \succ 0.$$

*Lemma 2:* Permuting the observations of an NND matrix preserves the NND structure.
*Proof:* Let $\pi$ be the permutation matrix such that it has exactly one entry in each row and in each column equal to 1, and all other entries are 0. For any permutation matrix, $\pi^{-1} = \pi^T$ and thus:

$$\pi \pi^T = \pi^T \pi = I$$

For any given NND matrix, $K$, $\pi K \pi^T$ is also NND. Clearly, $\pi K \pi^T$ is also symmetric as:

$$w^T (\pi K \pi^T) w = (\pi^T w)^T K (\pi^T w) \quad \forall w \neq 0 \quad \text{since } K \text{ is NND}$$

Furthermore, every NND matrix can be factored as $K = A^T A$, where $A$ is the Cholesky factor of $K$. The Cholesky factorization of NND matrices is numerically stable – a principal permutation of the rows and columns does not numerically destabilize the factorization (*30*). This leads to the result that symmetrically permuting the rows and columns of an NND matrix yields another NND matrix.

**Supplementary tables and figures**

**Table S1. List of technical indicators recorded for each observation/REIT by the CRSP Real Estate Database**

| Financial Indicator | Description |
| --- | --- |
| Rate of return on REIT stock | Total Return based on Used Prices and Used Dates and the CRSP distribution history. This is the variable to be classified by the SVM as positive or negative |
| Ordinary Dividends | Company's profits that get passed on to the shareholders |
| Capitalization | Market value of a company's outstanding shares |
| VW Return Index | A value-weighted (VW) stock market index whose components are weighted according to capitalization, and each stock in an index fund is not given the same importance. |
| Used Price | Combination of various good and soft prices used in an index according to index methodology rules. To be in an index, the security must be a valid security and have a good price, as well as an observed or soft price. |
| REIT Type | Description of whether the investment trust is of type mortgage, equity, or hybrid |
| Property Type | Description of whether the property type is residential, industrial, retail, healthcare or lodging/resorts. |
| Stock Exchange | Description of whether the securities were traded at NASDAQ, NY Stock Exchange (NYSE) or NYSE MKT. |

**Table S2. Test accuracy in forecasting whether the Rate of Return for an REIT would be positive or negative using SVMs for a range of kernel choices**

| Choice of Kernel | Training Accuracy | Testing Accuracy |
| --- | --- | --- |
| Laplacian | 74.4 | 67.7 |
| Radial Basis "Gaussian" | 68.7 | 67.3 |
| Semblance | 68.0 | 65.8 |
| Linear | 67.7 | 64.9 |
| Bessel | 67.5 | 66.6 |
| Polynomial | 66.9 | 68.9 |
| ANOVA | 60.9 | 56.3 |
| Hyperbolic tangent | 56.4 | 55.0 |
| Spline | 54.1 | 53.7 |

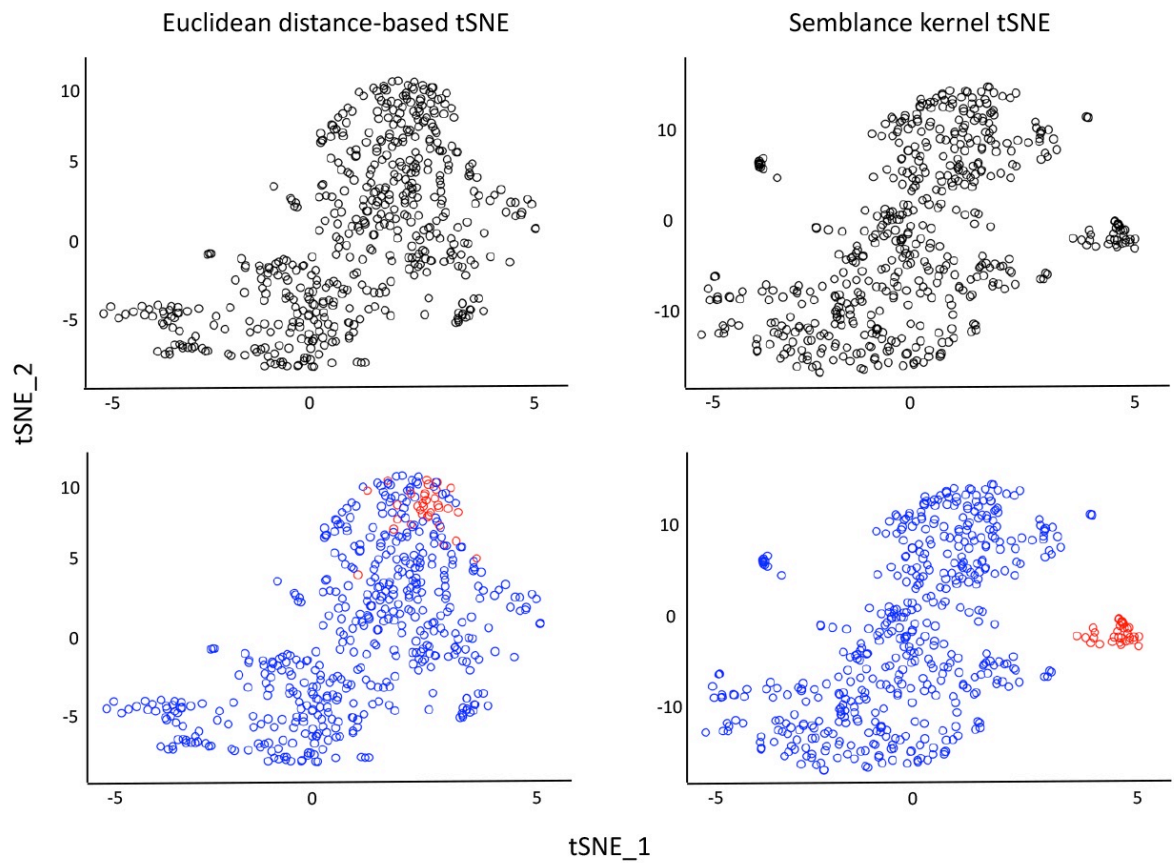

**Fig. S1.** We tested Semblance on an scRNA-seq dataset with 710 retinal horizontal cells (RHCs) (*16*), and compared its performance against the conventionally used, Euclidean distance-based analysis. The rank-based Semblance kernel leads to a better visual separation between two distinct RHC clusters, which were harder to distinguish from each other otherwise, based on Euclidean Distance.

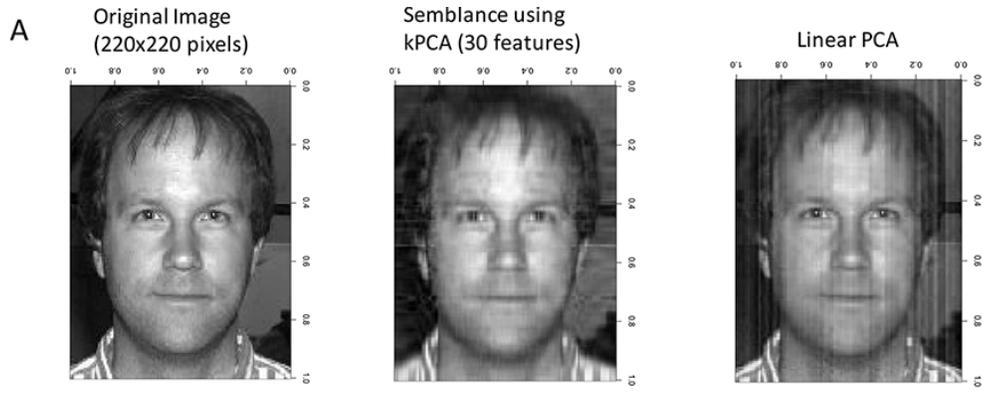

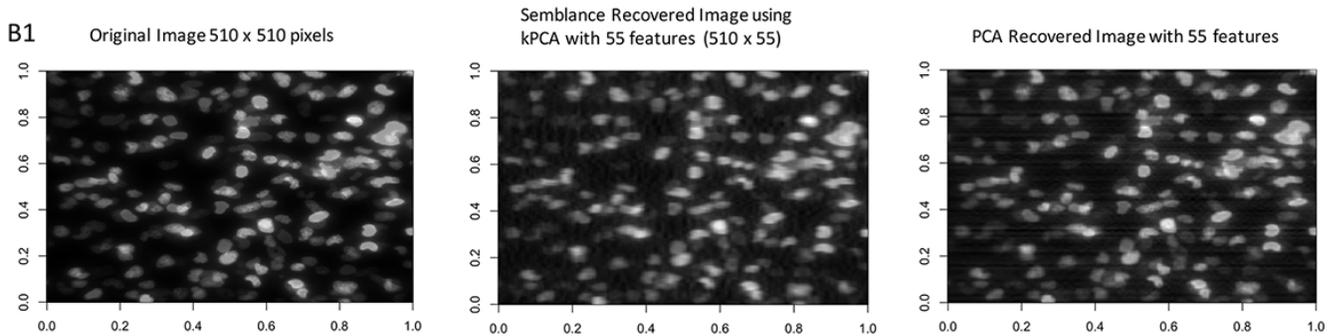

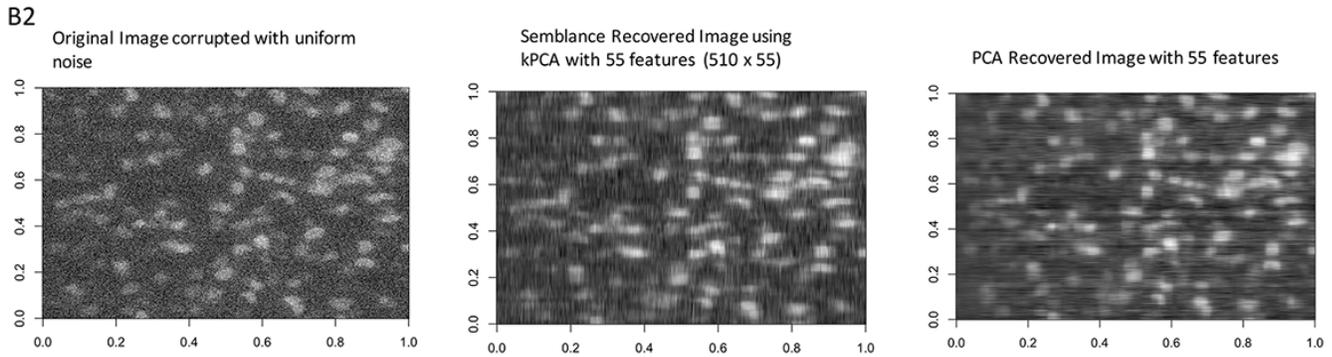

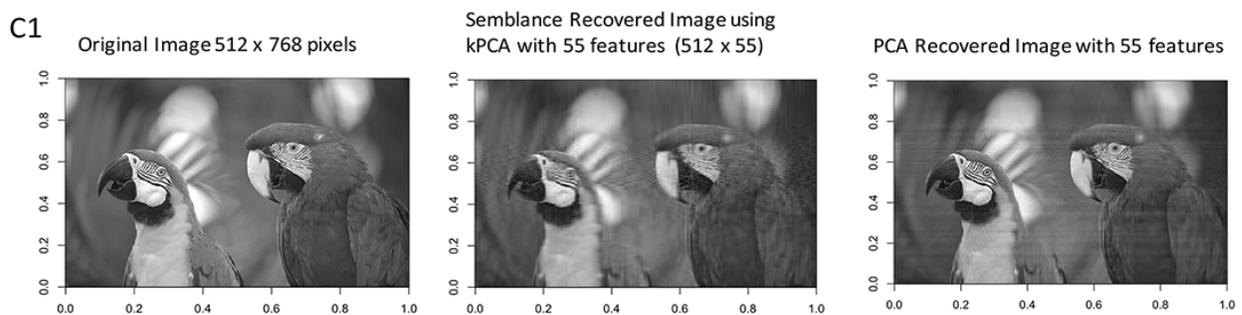

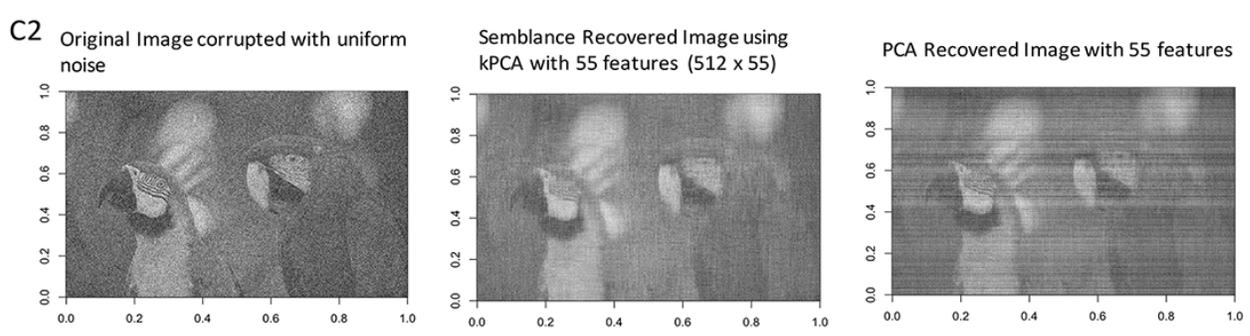

**Fig. S2.** kPCA using the Semblance kernel is able to efficiently compress and denoise images. Demonstrated examples are from the Yale Face Database **(A)**, and the EBImage package in *R* **(B-C)**. Panel B demonstrates an example of a microscopic image. Subpanels 1 in **(B-C)** show the results on the original images, whereas subpanels 2 display the results on corrupted images with added uniform noise. *Photo Credits:* EB Image (10.182129/B9.bioc.EBImage) and Yale Facebook Database B (http://vision.ucsd.edu/~iskwak/ExtYaleDatabase/ExtYaleB.html)